\documentclass{article}

\PassOptionsToPackage{numbers}{natbib}

\usepackage[final]{nips_2018}

\usepackage[utf8]{inputenc} % allow utf-8 input
\usepackage[T1]{fontenc}    % use 8-bit T1 fonts
\usepackage{hyperref}       % hyperlinks
\usepackage{url}            % simple URL typesetting
\usepackage{booktabs}       % professional-quality tables
\usepackage{amsfonts}       % blackboard math symbols
\usepackage{nicefrac}       % compact symbols for 1/2, etc.
\usepackage{microtype}      % microtypography
\usepackage{amsmath}  %I added this package
\usepackage{graphicx} %I added this package
\usepackage{multirow} %I added this package
\usepackage{siunitx}  %I added this package
\usepackage{setspace} % Lerroen arteko espazioa
\usepackage{fancyhdr} % Para cambiar las cabeceras de las pginas
\usepackage{pdfpages}

\usepackage{algorithmic}

\usepackage[ruled]{algorithm}

\usepackage[bottom]{footmisc}
\DeclareMathOperator*{\argmax}{argmax}
\DeclareMathOperator*{\argmin}{argmin}

\title{Universal adversarial examples in speech command classification}

\author{
  Jon Vadillo and Roberto Santana
     \\
  Department of Computer Science and Artificial Intelligence \\
  Faculty of Informatics\\
  University of the Basque Country\\
  \texttt{jvadillo005@ikasle.ehu.eus , rsantana@ehu.eus} \\
}

\providecommand{\keywords}[1]{\textbf{\textit{Keywords: }} #1}

\begin{document}

\cleardoublepage

\maketitle

\begin{abstract}
Adversarial examples are inputs intentionally perturbed with the aim of forcing a machine learning model to produce a wrong prediction, while the changes are not easily detectable by a human. Although this topic has been intensively studied in the image domain, classification tasks in the audio domain have received less attention. In this paper we address the existence of universal perturbations for speech command classification. We provide evidence that universal attacks can be generated for speech command classification tasks, which are able to generalize across different models to a significant extent. Additionally, a novel analytical framework is proposed for the evaluation of universal perturbations under different levels of universality, demonstrating that the feasibility of generating effective perturbations decreases as the universality level increases. Finally, we propose a more detailed and rigorous framework to measure the amount of distortion introduced by the perturbations, demonstrating that the methods employed by convention are not realistic in audio-based problems.

\end{abstract}


\section{Introduction}
\label{section::introduction}

The increasing applications of Deep Neural Networks (DNNs) in security-critical tasks, such as self-driving cars \cite{chen2015deepdriving}\cite{bojarski2016end} or voice controlled systems \cite{feng2017continuous}\cite{boles2017voice}\cite{gong2018overview}, require a high reliability on these computational models. In \cite{Szegedy:42503}, Szegedy et al. discovered that an attacker can fool a DNN with inputs which are slightly but adversarially modified: adversarial examples. These inputs are perturbed in such a way that the changes can not be detected by a human, while forcing a target model to misclassify or predict a specific output with high probability. The existence of such attacks questioned the reliability of DNNs in adversarial scenarios, and the reason for these vulnerabilities is still an open question \cite{gal2018idealised}.

The study of adversarial examples has focused primarily on the image domain and computer vision tasks \cite{akhtar2018threat}, whereas domains such as text or audio have received much less attention. Bearing in mind that the properties of adversarial examples are not well understood yet, studying the effect of these attacks in domains other than image-based tasks can provide very useful insights about the general properties of adversarial examples.

Due to the relevance of the vulnerability of DNNs to adversarial examples, in this paper we investigate some of the properties that are well known in image domain tasks, but, to the best of our knowledge, have not been studied yet in the audio domain. In particular, although methods that generate universal adversarial perturbations in the image domain have been proposed \cite{moosavi2017universal}\cite{mopuri2017fast}\cite{khrulkov2018art}, no such algorithms have been reported for many other tasks, such as speech classification in the audio domain. Thus, a better understanding of universal adversarial examples in different domains may provide useful insights into this property. 

The contributions of our work are the following: 1) we introduce a novel analytical framework for studying the existence of universal adversarial perturbations, 2) we provide evidence for the first time that universal perturbations can be generated in speech command classification tasks, and 3) we propose a more rigorous approach to evaluate the amount of distortion of the adversarial attacks in the audio domain, showing that the approaches used by convention are not realistic for audio problems. To facilitate future research, we made our code publicly available, along with the models and dataset used in this paper.\footnote{\url{https://github.com/vadel/AudioUniversalPerturbations}} 

While working on this paper, we were aware of research done by Neekhara \textit{et al.} \cite{neekhara2019universal} and Abdoli \textit{et al.} \cite{abdoli2019universal} which also addresses the generation of audio universal adversarial examples. There are important differences between these works and our contributions (discussed in Section \ref{sec:stateoftheart_universal}). 

The remainder of this paper is organized as follows. In the next section, we introduce the main concepts related to adversarial examples, as well as a novel taxonomy of universal adversarial perturbations. Section \ref{section:stateoftheart} provides an overview of the state-of-the-art in audio adversarial attacks, and Section \ref{section:attacks} describes previous attack algorithms used as the basis for our approach. Section \ref{section:instantiation} presents the selected dataset and models. The methodology used to craft audio universal perturbations is introduced in Section \ref{section::experimental_details}, and the experimental results are discussed in Section \ref{sec::universal_results}, focusing on different levels of the proposed taxonomy. Section \ref{section:instantiation_distortion} proposes a novel analytical framework for measuring the distortion introduced by audio perturbations. Section \ref{section::conclussion} concludes this paper and identifies lines for future research.

\section{Notation and assumptions}
\label{section:notation}
In this section we introduce the notation and present a number of assumptions used to derive our results.

\paragraph{Task}
We address adversarial perturbations in the audio domain for the particular problem of voice command classification. Specifically, the task of classifying an audio waveform among a finite set of possible (and previously defined) labels. We will assume audio waveforms of fixed lengths and composed of values in a bounded range.

\paragraph{Classification model} 
We will define a classification model $f$ as a function $f: X \rightarrow Y$, where $X\subseteq \mathbb{R}^d$ represents the $d$ dimensional input space of audio waveforms, and $Y$ the set of possible output labels, where $y_i\in \left\lbrace y_1 \dots y_k \right\rbrace$ represents the $i$-th label (commonly words or commands).

We will assume an end-to-end differentiable classification process, from which all the information will be available, so that the gradients of the model's output can be computed with respect to the input audio waveform. 

\paragraph{Adversarial example} 
An adversarial example $x'$ is defined as $x'=x+v$, where $x$ represents the original (clean) input, and $v \in \mathbb{R}^d$ represents the adversarial perturbation. We assume that the original input $x$ must be a sample correctly classified by the model, and $f(x')\neq f(x)$. This restriction ensures that the perturbation is actually fooling the model. In addition, the perturbation will be measured by a distortion metric $\varphi(x,x')\rightarrow \mathbb{R}$, and will be restricted by a maximum threshold $\epsilon$, so that $\varphi(x,x')\leq \epsilon$. This assumption tries to ensure that $x'$ is as similar as possible to $x$.

Given a clean input sample $x$ and a target class $t\neq f(x)$, a targeted adversarial example consists of a perturbed sample $x'=x+v$ which satisfies $f(x')=t$. In contrast, untargeted adversarial examples only require the output label to not be the same as the original one, that is, $f(x')\neq f(x)$.

Individual adversarial attacks are based on unique perturbations, crafted specifically to be applied to one particular input sample, so the same perturbation is not expected to work with a different input. In contrast, universal adversarial attacks are based on perturbations that can be applied to any clean input sample, that is, perturbations which are able to fool the model independently of the input. In this work we will consider three different types of universal adversarial perturbations, according to the number of classes for which it is expected to work:

\begin{itemize}
\item Level 1 - \textit{Single-class} universal perturbations: perturbations expected to fool the target model only for inputs of one particular class.

\item Level 2 - \textit{Multiple-class} universal perturbations: perturbations expected to fool target model only for a particular subset of $N$ classes, with $1<N<k$.

\item Level 3 - \textit{Fully} universal perturbations: perturbations expected to fool a target model for every possible input, with complete independence of the original class.
\end{itemize}

Notice that the three levels are particular cases of an abstract type of $N$-class adversarial perturbations, where $1\leq N\leq k$. However, for the sake of clarity we use this separation in levels to develop our experiments and analysis.

\section{State of the art}
\label{section:stateoftheart}
\subsection{Previous attacks in the audio domain}

The main difficulty of crafting adversarial examples in the audio domain is that, unlike the image domain, gradient-based methods are not directly applicable. This is due to the non-differentiability of the Mel-Frequency Cepstrum (MFC) transformation 
%\cite{muda2010voice}
%\cite{seki2017deep}
\cite{zhang2019discriminative}, a widely-used feature extraction process for speech recognition tasks, which is considered a standard component of speech recognition models. In addition, many error metrics employed in audio domain tasks are non-differentiable as well, such as \textit{Word Error Rate} (WER) or \textit{Character Error Rate} (CER), preventing the use of gradient-based attack methods \cite{cisse2017houdini}\cite{zhou2018improving}.

In \cite{cisse2017houdini}, Cisse \textit{et al.} introduced Houdini, a surrogate differentiable loss function which allows the use of  gradient-based optimization methods in domains in which the task loss function is non-differentiable, which is the case of automatic speech recognition. They were able to generate both targeted and untargeted attacks against a model based on the architecture of DeepSpeech2 \cite{amodei2016deep}. However, they were able to construct audio
adversarial examples targeting only phonetically similar phrases. In addition, they proved the transferability of their attacks in a black-box scenario by fooling the Google Voice application with the adversarial examples generated for their local model.
 
As in \cite{cisse2017houdini}, in \cite{carlini2018audio} Carlini \textit{et al.} achieved targeted and untargeted attacks against DeepSpeech \cite{hannun2014deep} (Mozilla's implementation \cite{DeepSpeechMozilla}) in automatic speech recognition, using an optimization approach. In comparison to \cite{cisse2017houdini}, they achieved more powerful targeted attacks, being able to obtain any output phrase, whether phonetically similar or not, as well as hiding speech from being transcribed or forcing music to be transcribed as speech.

Finally, in \cite{alzantot2018did} Alzantot \textit{et al.} avoid the gradient based optimization by using genetic algorithms to craft adversarial examples against speech command classification tasks, using the same model structure and dataset used in this paper. Their method allowed them to generate adversarial examples in a black box scenario by just adding noise to the 8 least-significant bits in 16 bits-per-sample audio files.

All these works are based on individual adversarial attacks in the audio domain. In contrast, in this paper we demonstrate the existence of universal adversarial perturbations, that is, a single perturbation which is able to fool a target model for any input audio, which is a more general and complex attack approach.

\subsection{Universal adversarial examples}
\label{sec:stateoftheart_universal}
In \cite{moosavi2017universal}, a universal adversarial perturbation $v$ is crafted by sampling a subset of input instances $X$, iteratively computing the minimal local perturbation $\bigtriangleup v_i$ that sends each instance $x_i\in X$ to the decision boundary of the model, and then adding this local perturbation  to $v$. The method is described in more detail in Section \ref{section:attacks}.

Only recently, the existence of universal adversarial perturbations in the audio domain for the task of speech transcription has been addressed \cite{neekhara2019universal}. This task can be formalized as a mapping from a sequence of $L$ audio frames from the input space $X$ to a sequence of characters from the space $Y$, typically composed of \textit{a-z}, \textit{space} and an empty-character $\epsilon$: $f: X^L \rightarrow Y^L$.
In particular, the task addressed in \cite{neekhara2019universal} is the generation of a universal perturbation $v$ capable of producing an error rate $E(f(x), f(x+v))$ above a certain threshold $t$ for any input sample $x$. The metric $E(x,y)$ used to compute the produced error rate is a normalized Levenshtein distance \cite{yujian2007normalized}, that is, the percentage of edited characters using character removal, insertion or substitution operations. The approach used to generate the universal perturbations is the same as that proposed in \cite{moosavi2017universal}.

It is worth noting that speech command recognition from a predefined set of commands, the task addressed in this paper, has particular characteristics which are different to speech transcription. The main difference in the approach used to fool speech transcription models and speech command classification models might be the fact that, in transcription, there is an exponentially larger number of possibilities to increase the error rate.
On the contrary, in speech command classification, the whole signal is mapped to a single prediction, so that the possible outputs of the model are much more limited. The differences between these tasks imply that the feasibility of generating universal perturbations in one of them does not necessarily imply the same degree of feasibility in the other.

Similarly, the existence of universal perturbations for the sound classification task has been addressed recently in \cite{abdoli2019universal}. As in the previous case, this task is considerably different to the speech command classification task, as no specific component for speech treatment is involved. 

Furthermore, although these two works on audio universal perturbations make use of the same general approach that we employ to generate the attacks, they replace the method with which local perturbations are computed. In the original algorithm, introduced for the image domain, Deepfool \cite{moosavi2016deepfool} is used to construct the individual perturbations, which is based on pushing an input sample to the closest decision boundary. In \cite{neekhara2019universal}, an Iterative Gradient Sign Method \cite{kurakin2016adversarial} is employed, due to the non-tractability of Deepfool in speech transcription tasks, in which there exists decision boundaries for each audio frame across the time axis. In \cite{abdoli2019universal}, Deepfool is replaced by the Decoupled Direction and Norm attack \cite{rony2019decoupling}. On the contrary, we maintain the same algorithm proposed in the original approach, and introduce the required modifications for the method to work in the audio domain. 

Finally, we propose more rigorous standards to measure the amount of distortion introduced by the perturbations, demonstrating that the approaches employed by convention in previous works are not realistic in audio-based problems.

\section{Methods for generating adversarial perturbations}
\label{section:attacks}

To generate universal perturbations, in this paper we adapt the algorithm proposed by Moosavi-Dezfooli \textit{et al.} \cite{moosavi2017universal} for the image classification task (hereinafter referred to as UAP algorithm) to the audio domain.  This method relies on Deepfool \cite{moosavi2016deepfool}, a state-of-the-art algorithm to generate individual attacks. Therefore, in this section we briefly explain the main characteristics of both algorithms.

\subsection{Deepfool}

Deepfool \cite{moosavi2016deepfool} is an efficient greedy algorithm to generate untargeted individual adversarial examples, which iteratively approximates a perturbation that sends a sample $x$ to the decision boundary of the model. Let $x_0$ be an input of class $t$, $f$ a target classifier, and $f_j$ the output logits of $f$ corresponding to the class $j$. The region of space $\mathcal{R}$ in which $f$ predicts $t$ is defined as: 
\begin{equation}
\displaystyle \mathcal{R}=\bigcap_{j=1}^{k}\lbrace x: f_t(x)\geq f_j(x)\rbrace
\end{equation}
For general multiclass classifiers, at iteration $i$, the algorithm approximates the region $\mathcal{R}$ by a polyhedron $\tilde{\mathcal{R}_i}$:
\begin{equation}
\label{eq:deepfool_approximation_region}
\tilde{\mathcal{R}_i}=\bigcap_{j=1}^{k}\lbrace x: f_j(x_i) - f_t(x_i) + \bigtriangledown f_j(x_i)^\top x - \bigtriangledown f_t(x_i)^\top x \leq 0 \rbrace
\end{equation}
Thus, the distance between $x_i$ and the boundaries of the region ${\mathcal{R}}$ is approximated by the distance between $x_i$ and $\tilde{\mathcal{R}_i}$. Being $f'_j=f_j(x_i) - f_t(x_i)$ and  $w'_j=\bigtriangledown f_j(x_i) - \bigtriangledown f_t(x_i)$, the class $l\neq t$ for whose region is closest to $x_i$ is determined in the following way:\footnote{The $\ell_p$ norm used to measure the perturbations can be extended for any $p\in[1,\infty)$.}
\begin{equation}
\displaystyle  l =
\argmin_{j\neq t} \frac{|f'_j|}{||w'_j||_2}
\end{equation}
Finally, $x_i$ is updated by adding the following perturbation:
\begin{equation}
\label{eq:deepfool_update_rule}
\displaystyle  x_{i+1} \leftarrow x_i + (1+\epsilon)\frac{|f'_l|}{||w'_l||_2^2}w'_l,
\end{equation}
being $\epsilon\in \mathbb{R^+}$ an \textit{overshoot} parameter, used to increase the magnitude of the perturbation in order to surpass the decision boundary and reach a different decision region, fooling the model. This process is repeated until the perturbation changes the predicted class of $f$, that is, until the condition $\argmax_k f_k(x_i)\neq t$ is satisfied.

\subsection{UAP algorithm}
In \cite{moosavi2017universal}, an algorithm to find untargeted adversarial perturbations in the image domain is presented (hereinafter referred to as UAP), based on constructing a universal perturbation $v$ by aggregating individual perturbations generated using Deepfool. We will use this algorithm as a basis for our proposal.

Given a collection of input samples $X$, the algorithm iteratively computes the minimun local  perturbation $\bigtriangleup v_i$ that sends the image to the decision boundary of the model
\begin{equation}
\bigtriangleup v_i \leftarrow \argmin_r||r||_2 \text{ s.t. } f(x_i + v + r) \neq f(x_i)
\end{equation}
and then adds this local perturbation to $v$. 
In order to ensure that the norm of the perturbation is restricted $||v||_p<\xi$, after each modification, $v$ is projected on the $\ell_p$ ball of radius $\xi$ and centered at 0, using the following projection operator:
\begin{equation}
\mathcal{P}_{p,\xi}(v)=\argmin_{v'}||v-v'||_2 \text{ subject to } ||v'||_p \leq \xi
\end{equation}
Thus, at each iteration, the update rule of $v$ is defined as:
\begin{equation}
v \leftarrow \mathcal{P}_{p,\xi}(v + \bigtriangleup v_i)
\end{equation}

Once all the samples from $X$ have been processed, the fooling rate of the resulting perturbation $v$ is computed against $X$. This process is repeated until the empirical fooling rate of $v$ with respect to $f$ on the set $X$ is above a specified threshold $1-\alpha$:
\begin{equation}
\displaystyle \mathbb{P}_{x\in X}( \ f(x)\neq f(x+v) \ )>1-\alpha    
\end{equation}

\section{Speech Command Classification}
\label{section:instantiation}

The speech command classification problem consists of assigning a class to a given audio from a predefined set of $k$ classes. In order to introduce and validate the proposed approaches, we will use the Speech Command Dataset \cite{warden2018speech}, which consists of a set of WAV audio files of 30 different spoken commands. The duration of all the files is fixed to 1 second, and the sample-rate is 16kHz in all the samples, so that each audio waveform is composed of $16000$ values, in the range $[-2^{15},2^{15}]$. The original dataset contains samples of 30 different labels. We will use a subset of ten classes, the standard labels selected in previous publications \cite{alzantot2018did}\cite{warden2018speech}: "Yes", "No", "Up", "Down", "Left", "Right", "On", "Off", "Stop", and "Go". In addition to this set, we will consider two special classes: "Unknown" (a command different to the ones specified before) and "Silence" (no speech detected).

The target model that will be used to generate the adversarial examples is based on the CNN architecture described in \cite{Sainath2015convolutional}. This model takes as input a single WAV audio file (with a sample rate of 16kHz) and transforms the signal from the time domain to the frequency domain, from which a set of MFCC features are extracted for different time intervals. This results in a two dimensional channel, which is fed into the following topology: two convolutional layers with a rectified linear unit (ReLU) activation function, a fully-connected layer and a final softmax layer. 
The implementation has been built on the code provided in \cite{carlini2018audio} and \cite{tensorflow2015-whitepaper}, available from \footnote{\url{https://github.com/carlini/audio_adversarial_examples}} 
and 
\footnote{\url{https://github.com/tensorflow/tensorflow/tree/master/tensorflow/examples/speech_commands}} 
respectively. The model has been trained using the second version of the Speech Command Dataset \cite{warden2018speech}, achieving a Top-One Accuracy of $85.5$\% in the respective test dataset.\footnote{\url{http://download.tensorflow.org/data/speech_commands_test_set_v0.02.tar.gz}}

To analyze the effectiveness of the adversarial perturbations in a model other than the one used to generate them, we will use the pretrained speech recognition model available at \footnote{\url{http://download.tensorflow.org/models/speech_commands_v0.02.zip}}. Although this model has the same architecture as the one used to generate the attacks, it has different parameters and is based on a different computation of the MFCC features. This model achieves a Top-One accuracy of $82.5$\% on the same test dataset as Target Model A. We verified that there exist important differences between the predictions given by the models, as shown in Table \ref{initial_error_per_class}, in which the initial accuracy of each model is computed per class. In particular, the class \textit{Silence} obtains the highest accuracy in Target Model A but the lowest in Target Model B.

\begin{table}
  \centering
  \scalebox{0.85}{
  \begin{tabular}{|l|c|c|c|}
    %\toprule
    \hline
     Class & Target Model A & Target Model B  \\ \hline \hline
    %\midrule
    Silence & 99.51  & 56.61  \\ \hline
    Unknown & 66.42  & 71.08  \\ \hline
    Yes     & 94.03  & 91.17  \\ \hline
    No      & 74.57  & 82.47  \\ \hline
    Up      & 92.00  & 91.76  \\ \hline
    Down    & 80.79  & 79.31  \\ \hline
    Left    & 89.81  & 93.45  \\ \hline
    Right   & 88.64  & 90.66  \\ \hline
    On      & 87.12  & 87.63  \\ \hline
    Off     & 81.59  & 82.09  \\ \hline
    Stop    & 93.67  & 90.99  \\ \hline
    Go      & 77.36  & 72.14  \\ \hline
    %\bottomrule
  \end{tabular}
  }
    \caption{Initial accuracy per class of the target models}
  \label{initial_error_per_class}
\end{table}

\section{Methodology: Hill Climbing Reformulation of Universal Adversarial Perturbations}
\label{section::experimental_details}
To generate universal perturbations, we use Algorithm \ref{algorithm_universal_stable} (hereinafter referred to as UAP-HC), which is a reformulation of the UAP algorithm proposed for images \cite{moosavi2017universal}, adapting it to the audio domain. An additional change is incorporated to the algorithm: the perturbation is updated in a \textit{Hill-Climbing} fashion only if the update improves the global fooling rate of $v$, which is determined by the ratio of fooled samples of $X$ when applying a perturbation (line 12). The change was motivated because, after an initial experimentation, the obtained results revealed that updating the universal perturbation $v$ with every new local perturbation $\bigtriangleup v_i$ (assuming that $f(x_i)=f(x_i+v)\neq f(x_i+v+\bigtriangleup v_i)$, being $x_i$ the input sample for which $\bigtriangleup v_i$ is crafted) led to a highly unstable evolution of the fooling rate. This modification may not be required for other audio tasks, since in the research reported in \cite{neekhara2019universal} and \cite{abdoli2019universal} the update rule is maintained without changes.

\begin{algorithm}[!t]
 \caption{UAP-HC}
  \label{algorithm_universal_stable}
\begin{algorithmic}[1]
 \REQUIRE { A classification model $f$,  a set of input samples $X$, a projection operator $\mathcal{P}_{p,\xi}$ and a fooling rate threshold $\alpha$}
 \ENSURE{ A universal perturbation $v$}
 
 \STATE $v \leftarrow $ initialize with zeros
 \STATE $foolingRate \leftarrow$ 0
 \WHILE{$foolingRate < $ $1-\alpha$}
  \FOR{$x_i \in X$}
  	  %\tcp{Check that $x_i$ is not already fooled by $v$}
  	  \STATE Check that $x_i$ is not already fooled by $v$
  	  \IF{$f(x_i) = f(x_i+v)$}
	    %\tcp{Compute local perturbation using Deepfool algorithm}
	    \STATE Compute local perturbation using Deepfool
	  	\STATE $\bigtriangleup v_i \leftarrow$ deepfool($x_i + v$)\\
	  	%\tcp{Update $v$ only if adding the local perturbation $\bigtriangleup v_i$ increases the global fooling ratio.}
	  	\STATE Update $v$ only if adding $\bigtriangleup v_i$ increases the global fooling rate. 
	  	\STATE $v' \leftarrow \mathcal{P}_{p,\xi}(v+\bigtriangleup v_i)$ 
	  	\STATE $foolingRate' \leftarrow \displaystyle \mathbb{P}_{x\in X}( \ f(x)\neq f(x+v') \ )$
	  	\IF{$foolingRate < foolingRate'$}
			\STATE $v \leftarrow v'$\\	 
  			\STATE $foolingRate \leftarrow foolingRate'$
  			%$fooling\_rate \leftarrow \displaystyle \mathbb{P}_{x\in X}( \ f(x)\neq f(x+v) \ )$\\			 		
	  	\ENDIF	 
	 \ENDIF  
  \ENDFOR
 \ENDWHILE
\end{algorithmic}
\end{algorithm}

Regarding the parameters of the algorithm, the universal perturbation is bounded by the $\ell_2$ norm, with a threshold value of $\xi=0.1$. In addition, the maximum number of iterations allowed for the Deepfool algorithm is set to $100$. The overshoot parameter of the Deepfool algorithm is set to $0.1$.\footnote{The parameters of the Deepfool algorithm were selected after a preliminary experiment on individual adversarial examples. The selected parameters managed to fool the model for 99.9\% of the inputs that we evaluated.} Finally, the fooling rate threshold of UAP-HC is set to $0.9$ ($\alpha=0.1$), and the algorithm is restricted to 5 iterations, that is, 5 complete passes through the entire set $X$.

The effectiveness of the perturbation in both the subset of samples used to craft the perturbation (training set) and the subset used to measure its effectiveness on unseen samples (validation set)  is computed as the percentage of originally correctly classified samples for which the prediction of the model changes after the adversarial attack. We will refer to this metric as \textit{fooling ratio} (\textit{FR}) \cite{moosavi2017universal}.  

In all the experiments, the set used to craft the perturbations is composed of $N$ samples randomly selected from the same training dataset used to train our target model (without replacement). In all the cases, 5 different trials are made, using different samples of the training set, and the results are reported for the perturbation that maximizes the fooling ratio on the training set. The validation set is the one used to test the initial accuracies of the target models in Section \ref{section:instantiation}.

Finally, following the approach employed in previous works on adversarial examples in speech related tasks \cite{neekhara2019universal}\cite{carlini2018audio}, the relative loudness of the adversarial perturbation $v$ with respect to an audio sample $x$ is measured according to the following metrics, in Decibels (dB): $dB_x(v)=dB(v)-dB(x)$, where $dB(x)=\max_{i} \ 20\cdot log_{10}(|x_i|)$. As the loudness of the perturbation should be lower than the loudness of the original signal, the difference is expected to be below zero. A distortion level of -32dB is established as the maximum acceptable distortion, which roughly corresponds to the difference between ambient noise in a quiet room and a person talking \cite{carlini2018audio}\cite{smith1997scientist}. In our results, we report the mean relative loudness on the validation set to measure the distortion produced by the universal perturbation generated.

\section{Results}
\label{sec::universal_results}

\subsection{Level 1 perturbations}
We initially focused on the lowest universality level according to the taxonomy introduced, in which we attempt to generate a single perturbation which is able to fool any input of one particular class. This analysis may highlight the robustness or weakness of different classes against universal perturbations.

The fooling ratio in both training and validation sets and the mean distortion level of the perturbations are presented in Table \ref{tab:level1_results} for each class. According to the results, the effectiveness of the algorithm is remarkably high, also requiring a considerably low amount of distortion, far below the threshold of -32dB except for class \textit{silence}\footnote{This effect on class \textit{silence} is due to the fact that, due to the nature of the samples corresponding to that class, their loudness level is lower than for the rest of classes.}. For 7 out of 12 classes the best perturbation fools at least 70\% of the samples in the validation set, and for 10 out of 12 classes it fools at least 60\%. Clearly, generating universal perturbations is more difficult in some classes than in others, since the fooling ratios obtained are of a considerably different magnitude. Although some of the classes obtain fooling ratios above 80\%, as is the case of \textit{no}, others such as \textit{silence}, \textit{left} or \textit{stop} are below 62\%. In particular, in the case of the class \textit{silence}, the fooling ratio is below 25\% in both training and validation sets.

\begin{table}[t]
\centering
\scalebox{0.85}{
\begin{tabular}{|c||S[table-format=2.2]|S[table-format=2.2]||S[table-format=2.2]|S[table-format=2.2]||c|}
\hline
\multirow{2}{*}{Class} & \multicolumn{2}{c||}{FR\% Target Model A} & \multicolumn{2}{c||}{FR\% Target Model B} & \multicolumn{1}{l|}{\multirow{2}{*}{Mean dB}} \\ \cline{2-5}
 & {\ \ Train \ \ } & {Valid} & {\ \ Train \ \ } & \multicolumn{1}{c||}{Valid} & \multicolumn{1}{c|}{} \\ \hline \hline
Silence & 23.80 & 19.46 & 12.96 & 12.55 & -27.26 \\ \hline
Unknown & 72.70 & 73.06 & 21.30 & 31.38 & -39.64 \\ \hline
Yes     & 74.50 & 74.36 & 62.58 & 67.02 & -41.72 \\ \hline
No      & 86.50 & 83.77 &  4.38 &  7.19 & -45.04 \\ \hline
Up      & 84.20 & 75.45 &  9.17 &  7.95 & -49.15 \\ \hline
Down    & 71.50 & 65.55 & 14.15 & 12.73 & -41.07 \\ \hline
Left    & 52.30 & 49.73 & 13.40 & 17.66 & -38.43 \\ \hline
Right   & 68.70 & 63.82 & 12.29 & 15.04 & -43.48 \\ \hline
On      & 76.00 & 75.65 & 7.55  &  9.22 & -42.75 \\ \hline
Off     & 80.10 & 73.48 & 26.81 & 37.58 & -43.73 \\ \hline
Stop    & 61.40 & 61.82 & 19.77 & 25.40 & -44.65 \\ \hline
Go      & 87.80 & 80.06 & 29.78 & 32.41 & -42.62 \\ \hline
\end{tabular}
}
\caption{Results obtained in Level 1 universal perturbations}
\label{tab:level1_results}
\end{table}

\subsection{Level 2 perturbations}
As explained in Section \ref{section:notation}, Level 2 universal perturbations attempt to create a single perturbation which is able to fool all the inputs of a particular subset of classes. Their study allows us to investigate how the difficulty of the problem changes among different subsets of classes, helping to identify those more vulnerable to universal perturbations. It also opens the question of whether Level 2 perturbations are less or more effective than Level 3 perturbations. Such insights could be used in future research as a basis to study different properties of universal attacks and to obtain a better understanding of these vulnerabilities.

The experimentation initially focuses on subsets of two classes. For illustration purposes, we selected a reduced and diverse set of pairs in terms of the phonetic similarity and transition probabilities between classes. The following pairs are considered:  \textit{no-go}, \textit{up-off}, \textit{yes-left}, \textit{unknown-stop}, \textit{silence-unknown}, and \textit{no-up}. The training set is composed of 500 samples per class, forming a total of 1000 training samples.

Results are shown in Table \ref{tab:level2_results}. For each pair of classes, the fooling ratio has been computed considering the audio samples of the two classes involved (\textit{Total}) as well as individually in each class (\textit{C1} and \textit{C2}). The mean distortion levels of the perturbations are also displayed in the table, computed from the validation set. According to the results, for the pairs \textit{no-go}, \textit{up-off} and \textit{no-up} we obtained remarkably high fooling ratios, being also the most balanced pairs regarding the values obtained at each class. On the contrary, for the remaining pairs we obtained a final fooling ratio below 50\%, being also highly unbalanced on the ratios per class, with the exception of the pair \textit{unknown-stop}, in which the difference between both classes is approximately 13\% in the training set and approximately 5\% in the validation set.

\begin{table}[h]
\centering
\scalebox{0.73}{
\begin{tabular}{|l||S[table-format=2.2]|S[table-format=2.2]|S[table-format=2.2]||S[table-format=2.2]|S[table-format=2.2]|S[table-format=2.2]||S[table-format=2.2]|S[table-format=2.2]|S[table-format=2.2]||S[table-format=2.2]|S[table-format=2.2]|S[table-format=2.2]||c|}
\hline
\multirow{3}{*}{Pair} & \multicolumn{6}{c||}{FR\% - Target Model A} & \multicolumn{6}{c||}{FR\% - Target Model B} & \multirow{3}{*}{Mean dB} \\ \cline{2-13}
 & \multicolumn{3}{c||}{Train} & \multicolumn{3}{c||}{Valid} & \multicolumn{3}{c||}{Train} & \multicolumn{3}{c||}{Valid} &  \\ \cline{2-13}
 & {Total} & {C1} & {C2} & {Total} & {C1} & {C2} & {Total} & {C1} & {C2} & {Total} & {C1} & {C2} &  \\ \hline \hline
\begin{tabular}[c]{@{}l@{}}C1: no\\ C2: go\end{tabular} & 71.20 & 68.80 & 73.60 & 68.35 & 72.52 & 64.31 & 41.36 & 37.50 & 45.52 & 49.04 & 47.90 & 50.34 & -46.37 \\ \hline
\begin{tabular}[c]{@{}l@{}}C1: up\\ C2: off\end{tabular} & 67.50 & 70.40 & 64.00 & 62.03 & 62.40 & 61.58 & 28.14 & 31.70 & 24.69 & 31.94 & 34.10 & 29.39 & -45.31 \\ \hline
\begin{tabular}[c]{@{}l@{}}C1: yes\\ C2: left\end{tabular} & 36.90 & 73.80 & 0.00 & 36.78 & 71.32 & 0.00 & 29.12 & 58.79 & 0.00 & 32.20 & 64.60 & 0.00 & -38.83 \\ \hline
\begin{tabular}[c]{@{}l@{}}C1: unk.\\ C2: stop\end{tabular} & 48.90 & 55.20 & 42.60 & 47.41 & 50.55 & 45.19 & 15.21 & 10.57 & 19.20 & 21.54 & 16.90 & 25.13 & -41.16 \\ \hline
\begin{tabular}[c]{@{}l@{}}C1: silence\\ C2: unk. \end{tabular} & 43.30 & 19.40 & 67.20 & 35.60 & 16.75 & 63.84 & 27.56 & 24.65 & 29.63 & 32.25 & 21.65 & 40.69 & -36.32 \\ \hline
\begin{tabular}[c]{@{}l@{}}C1: no\\ C2: up\end{tabular} & 70.90 & 71.60 & 70.20 & 65.95 & 70.86 & 62.15 & 7.86 & 3.97 & 11.79 & 9.12 & 6.89 & 11.03 & -42.04 \\ \hline
\end{tabular}
}
\caption{Results obtained in Level 2 Universal Perturbations}
\label{tab:level2_results}
\end{table}

\subsection{Level 3 perturbations}
Level 3 universal perturbations comprise the most general level of universality, in which we attempt to create a single \textit{input agnostic} perturbation which is able to fool any possible input. This particular level is the only one investigated in all previous works on universal adversarial examples.

Results for Level 3 perturbations are shown in Table \ref{tab:level3_results}. Comparing the fooling ratio achieved in each class, it can be seen that the perturbation effectively fools a considerable percentage of samples in the majority of the classes. In particular, in 2 out of 12 classes the fooling ratio is at least 70\% in the training set, and for 8 out of 12 classes at least 60\%. In the validation set, the fooling ratio is at least 60\% for 6 classes. As for previous levels, the mean distortion is below -32dB in all the cases except for the class \textit{silence}.

However, the total fooling ratio is reduced because of the low results obtained in classes \textit{silence} and \textit{left}, achieving a total final fooling ratio near 50\% in both training and validation sets. Although room for improvement exists regarding the effectiveness of the perturbations, it is noteworthy that the attacks generated imply a noticeable drop of the accuracy of the model, of approximately 50\% in both sets. Furthermore, the high fooling rates obtained in the majority of the classes proves the existence of universal perturbations in speech command classification.

\begin{table}[h]
\centering
\scalebox{0.85}{
\begin{tabular}{|c||S[table-format=2.2]|S[table-format=2.2]||S[table-format=2.2]|S[table-format=2.2]||c|}
\hline
\multirow{2}{*}{Class} & \multicolumn{2}{c||}{FR\% Target Model A} & \multicolumn{2}{c||}{FR\% Target Model B} & \multirow{2}{*}{Mean dB} \\ \cline{2-5}
 &  {\ \ Train \ \ } & {Valid} & {\ \ Train \ \ } & {Valid} &  \\ \hline \hline
Silence  & 26.00 & 18.72 & 36.07 & 35.50 & -23.88 \\ \hline
Unknown  & 62.00 & 61.62  & 44.30 & 49.31 & -36.66 \\ \hline
Yes      & 62.00 & 66.24  & 63.33 & 64.14 & -37.56 \\ \hline
No       & 61.00 & 66.23  & 44.21 & 51.20 & -36.74 \\ \hline
Up       & 70.00 & 61.89  & 52.08 & 50.26 & -38.18 \\ \hline
Down     & 52.00 & 44.82  & 22.58 & 15.53 & -37.97 \\ \hline
Left     & 0.00  &  0.00  &  0.00 &  0.26 & -37.60 \\ \hline
Right    & 60.00 & 55.84  & 35.05 & 38.99 & -37.58 \\ \hline
On       & 60.00 & 57.39  & 26.88 & 28.24 & -36.42 \\ \hline
Off      & 73.00 & 68.60  & 43.16 & 48.48 & -38.27 \\ \hline
Stop     & 45.00 & 43.38  & 29.17 & 33.42 & -38.25 \\ \hline
Go       & 60.00 & 63.34  & 44.57 & 51.72 & -38.01 \\ \hline
Total    & 52.58 & 46.64  & 36.53 & 38.70 & -36.43 \\ \hline
\end{tabular}
}
\caption{Results obtained in Level 3 Universal Perturbations}
\label{tab:level3_results}
\end{table}

\subsection{Transferability}
The transferability levels of the generated universal perturbations is described in Tables  \ref{tab:level1_results}, \ref{tab:level2_results} and \ref{tab:level3_results}, for Level 1, Level 2 and Level 3 universal perturbations respectively. Similarly to the original target model, the effectiveness of the perturbations is considerably higher for some classes or subsets of classes than for others. In Level 1 perturbations, for 4 out of 12 classes the fooling ratio is above 30\% in the validation set. In Level 2 perturbations, for 4 out of 6 pairs of classes the fooling ratio is above 30\%. Finally, for Level 3 perturbations, although the total fooling ratio in the validation set is 38.70\%, for 4 out of 12 classes the fooling ratio is above 50\%, and for 9 out of 12 classes above 30\%.

As can be seen, the generalization of the perturbations across models is generally higher as the universality level increases, which may reflect that, when a large number of classes are considered, the perturbations are able to generalize better not only across classes, but also across target models.

\subsection{Discussion}
In this section we provide evidence that universal adversarial examples can be created for speech command classification models, although room for improvement exists regarding the effectiveness of the attacks. In addition, although the effectiveness of the perturbations clearly decreases when they are transfered to a new target model, the results indicate that they are transferable to a significant extent.

The analysis carried out for different levels of universality revealed some interesting insights about universal perturbations in this domain: 1) it is considerably easier to create perturbations for some subsets of classes than for others, 2) while generating Level 1 universal perturbations is considerably more feasible than Level 2 or Level 3 perturbations, generating Level 3 perturbations is not necessarily more difficult than Level 2 perturbations and 3) the transferability of the attacks increases as the universality level increases. All these conclusions could be used as a basis to better study the robustness of the model against adversarial attacks or to provide a theoretical explanation of universal perturbations.

\section{Deeper evaluation of the amount of distortion}
\label{section:instantiation_distortion}
As previously explained, to measure the loudness of an audio signal $x=\left\lbrace x_1,...,x_d \right\rbrace$ of length $d$, previous related works \cite{neekhara2019universal}\cite{carlini2018audio} make use of the maximum decibel level of the signal. We have shown in Section \ref{sec::universal_results} that, under this criterion, our perturbations were far below the threshold of -32dB, and therefore, assumed to be quasi-imperceptible according to the standards established by convention. 

The drawback of considering only the maximum decibel level of the signals to compute the distortion is that this metric can be highly sensitive to peaks in the intensity of the audio, which may lead to results that do not adequately represent the overall loudness level of the audio. To obtain a more accurate and robust evaluation of adversarial examples, we propose using, in addition to this distortion metric, another one based on the mean loudness of the signal:
\begin{equation}
\label{eq:mean_metric}
dB_{mean}(x) = 20\cdot log_{10}\left(\frac{1}{d}\sum_{i=1}^d{|x_i|}\right)
\end{equation}
As in the previous case, the distortion level of a signal $v$ with respect to another reference signal $x$ is measured as:
\begin{equation}
\label{eq:mean_metric_dif}
dB_{x,mean}(v) = dB_{mean}(v)-dB_{mean}(x).
\end{equation}

Besides, previous works compute the distortion between two signals by applying the metrics to the entire signals. In order to provide a more detailed evaluation of the distortion, we propose analyzing the perturbation for different parts of the signals independently, partitioned according to the information they convey. In particular, we will consider two different parts in each audio signal: the \textit{vocal} part and the \textit{background} part. As both parts are significantly different, and because the distortion will be measured based on aggregated features of the audio signals, evaluating these metrics in the entire signal could lead to misleading results. Thus, we will study the distortion in these two parts independently, in order to obtain more informative results. To the best of our knowledge, a similar type of analysis has not been previously reported in the related literature.

\begin{figure}[h]
\centering
\includegraphics[scale=0.5]{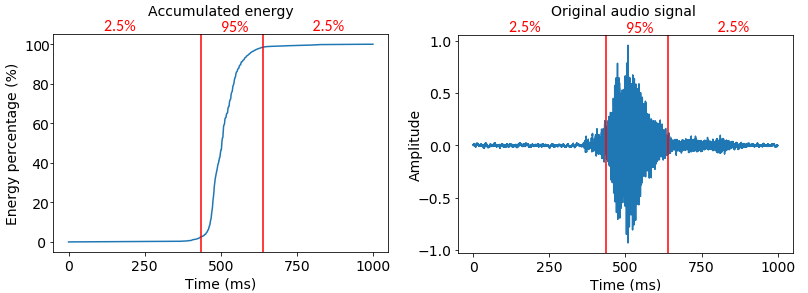}
\caption[Illustration of the identification of the vocal part in an audio signal according to the cumulative sum of its energy.]{Illustration of the identification of the vocal part in an audio signal (right) according to the cumulative sum of its energy (left). Red vertical lines delimit the vocal part, and the red numbers located at the top of each image indicate the percentage of energy located in each interval.}
\label{example_vocal_part}
\end{figure}

Being $x=\left\lbrace x_1...x_d \right\rbrace$ an audio signal, the vocal part will be delimited by the continuous range $[x_a,x_b]$ containing 95\% of the energy of the signal, that is, $\frac{\sum_{j=a}^{b}x_j^2}{\sum_{i=1}^{d}x_i^2}\approx 0.95$. Thus, we will assume that ranges $[x_1,x_a)\cup (x_b,x_d]$ will be composed just of background noise. Figure \ref{example_vocal_part} shows an audio signal in which the vocal part has been delimited using this method. Notice that this partition is well suited for single command audios in which it is assumed that the vocal part of the signal is contiguous.

\begin{figure}[!h]
\centering
\includegraphics[scale=0.4]{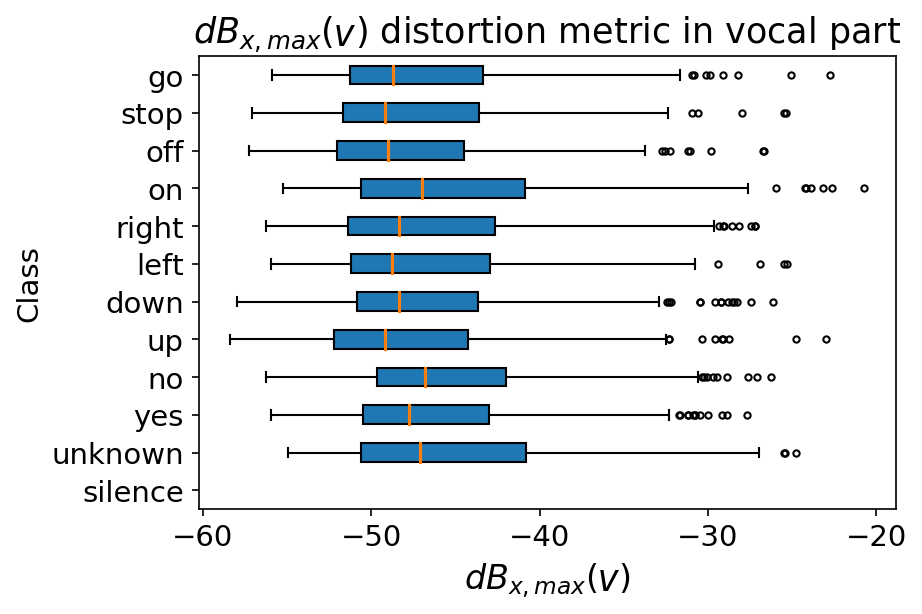}
\includegraphics[scale=0.4]{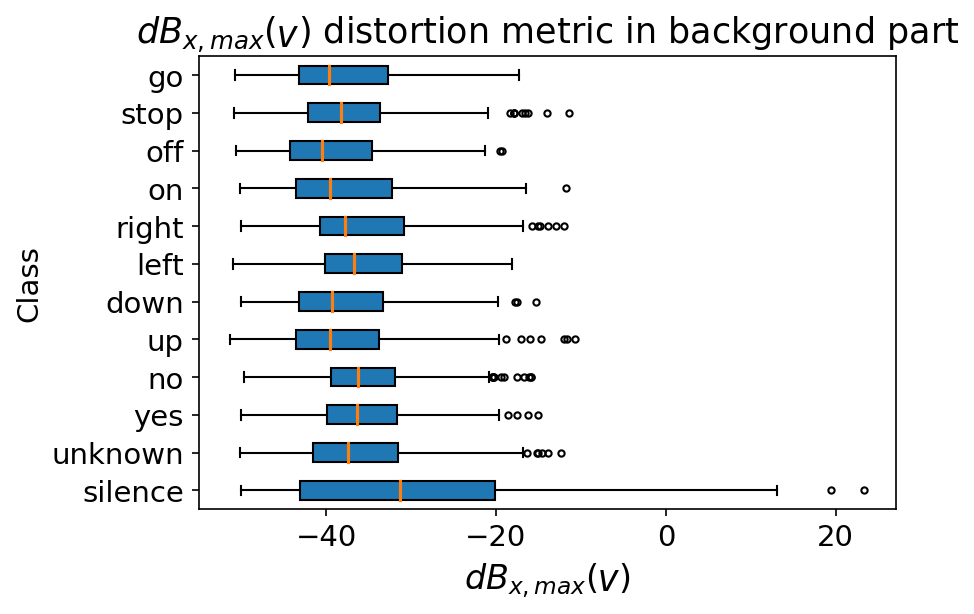}
\includegraphics[scale=0.4]{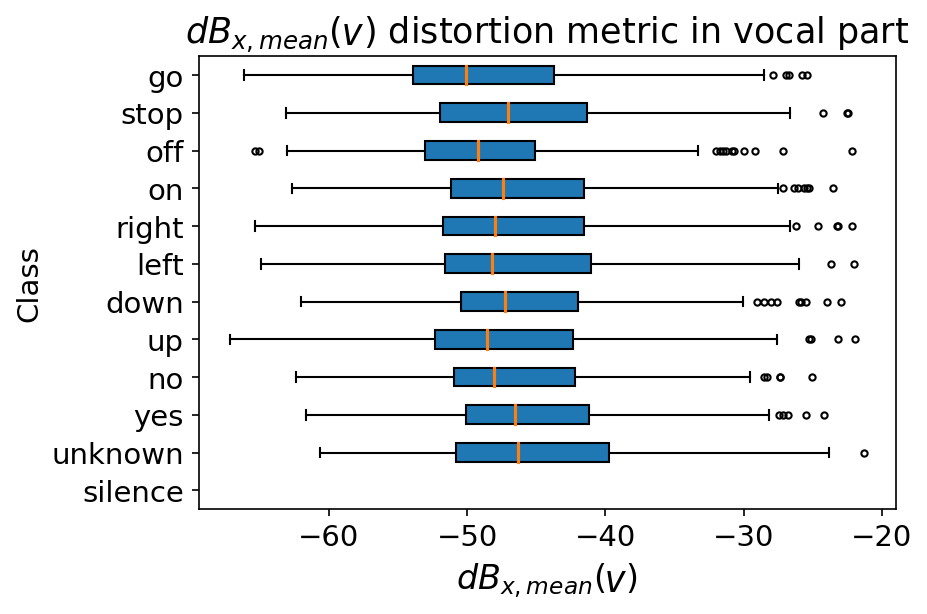}
\includegraphics[scale=0.4]{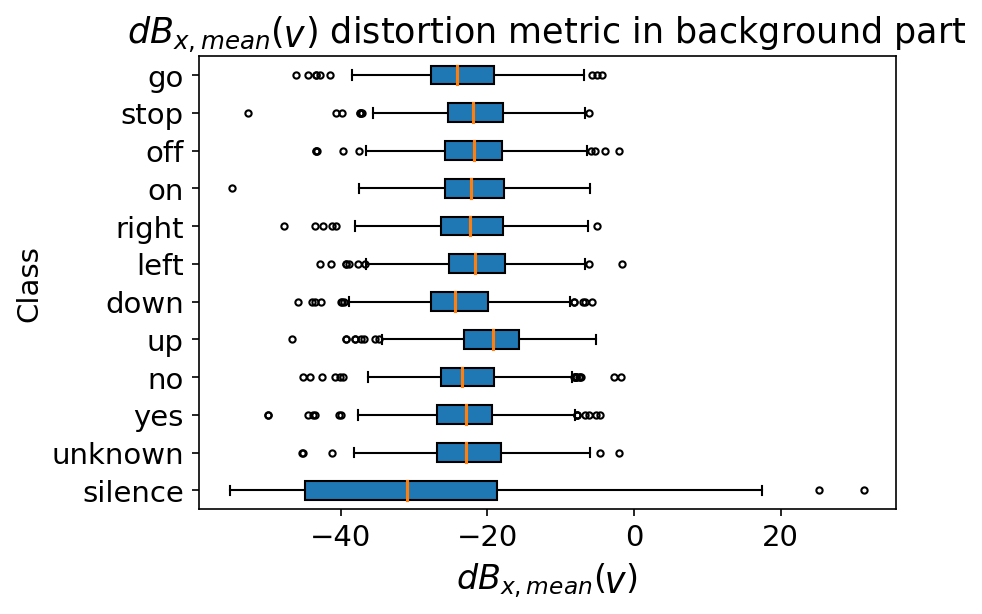}
\caption[Distortion level of the generated Level 3 universal perturbations, evaluated in the validation set.]{Distortion level of the generated Level 3 universal perturbations, evaluated in the validation set using $dB_{x,max}(v)$ metric (top row) and $dB_{x,mean}(v)$ metric (bottom row). For each audio, the distortion has been measured in the background part as well as in the vocal part, except for the class silence, as it is only composed of background noise. Results are averaged for the 5 experiments carried out.}
\label{fig:universal3_distortion}
\end{figure}

The amount of distortion computed using this approach is presented in Figure \ref{fig:universal3_distortion}, represented for the Level 3 universal perturbations. According to the results, the amount of distortion in the vocal part is acceptable for the great majority of the samples for the two metrics considered. However, the distortion in the background part is above the threshold of $-32$dB for a considerably high proportion of samples according to the metric $dB_{x,mean}(v)$, which indicates that the perturbation may be highly detectable outside the vocal part. We encourage the reader to listen to some adversarial examples, to empirically verify that the perturbations are actually detectable.\footnote{\url{https://vadel.github.io/audio_adversarial/UniversalPerturbations.html}} These results strengthen our proposal about the need to employ more rigorous approaches in order to measure and threshold the distortion produced by the adversarial perturbations in a more representative way. 

\section{Conclusions}
\label{section::conclussion}

In this work the feasibility of generating universal adversarial perturbations in the task of speech command classification has been addressed. To deal with this question, we have introduced the concept of $N$-class universality as a theoretical framework for the study of universal perturbations, which is based on the number of target classes that the universal attack is designed to inhibit. This framework has been used to determine the feasibility of generating universal perturbations according to different levels of universality.

The reported results demonstrate that effective universal adversarial attacks can be generated for speech command classification models, although improvement is needed in order to consider the perturbations fully universal. The results also reflected that, although the effectiveness of the perturbations was reduced in new target models, they were transferable across models to a significant extent.

Regarding the evaluation of the detectability of the attacks, we have proposed the use of an experimental framework to perform a deeper evaluation of the distortion in the audio domain, based on measuring the distortion in different parts of the audio signals, partitioned according to the information they convey. We have shown that, although the obtained distortion levels are acceptable according to the standard evaluation approach employed by convention in previous works, under a more rigorous evaluation, the perturbations were detectable outside the vocal parts corresponding to the speech command.  This fact should be addressed in future research to achieve less detectable perturbations.

Overall, we have provided evidence that universal adversarial attacks can be generated for speech command classification models, which supposes a serious concern regarding the reliability of speech-based agents in practical scenarios.

The performed study also leaves a number of interesting open questions and future research lines. First of all, the audio domain contains a number of properties that could be used to enhance the universal perturbations, for example, the frequency domain. A wiser exploitation of such properties may lead to more powerful attacks or produce less detectable perturbations. In addition, the reason of the DNN vulnerabilities to adversarial examples is still an open question, although many different hypotheses have been proposed \cite{goodfellow2014explaining}\cite{schmidt2018adversarially}\cite{ilyas2019adversarial}. An evaluation of those hypotheses in the studied domain may be an interesting research line, in order to address their validity. 

Finally, previous works on universal adversarial perturbations in the image domain revealed the existence of a low dimensional subspace containing vectors with high attack capabilities \cite{moosavi2017universal}\cite{moosavi2017analysis}. Finding such subspaces in the audio domain would be an important discovery, which may provide valuable insights from a theoretical point of view, as it may provide advantages regarding the connection between different domains.

\section*{Acknowledgment}

The authors would like to thank to the Intelligent Systems Group (University of the Basque Country, Spain) for providing the computational resources needed to develop the project. Roberto Santana acknowledges support by the Basque Government (IT1244-19 and ELKARTEK programs), and  Spanish Ministry of Economy and Competitiveness MINECO (project TIN2016-78365-R).

\bibliographystyle{unsrt}
\bibliography{references}

\vfill

\end{document}